\documentclass{article}
\usepackage{ijcai24}

\usepackage{times}
\usepackage{soul}
\usepackage{url}
\usepackage[hidelinks]{hyperref}
\usepackage[utf8]{inputenc}
\usepackage[small]{caption}
\usepackage{graphicx}
\usepackage{amsmath}
\usepackage{amsthm}
\usepackage{booktabs}
\usepackage{subcaption}
\usepackage{algorithm}
\usepackage{algorithmic}
\usepackage[switch]{lineno}
\usepackage{amsfonts,amssymb}
\usepackage{multirow}
\usepackage{hhline}
\usepackage{diagbox}
\usepackage{xcolor}
\usepackage{adjustbox}
\usepackage{tabularx}
\usepackage{booktabs}
\usepackage{bigstrut}
\usepackage{rotating}
\usepackage{makecell}

\title{Strengthening Layer Interaction via Dynamic Layer Attention}

\author{
Kaishen Wang$^1$\and
Xun Xia $^{2}$\and
Jian Liu$^2$\and
Zhang Yi$^1$\and
Tao He$^{1,}$\footnote{Corresponding author.}
\affiliations
$^1$College of Computer Science, Sichuan University \\
$^2$Clinical Medical College and The First Affiliated Hospital of Chengdu Medical College
\emails
wangks@stu.scu.edu.cn, xiaxun@cmc.edu.cn, liujiansh@126.com,
\{zhangyi, tao\_he\}@scu.edu.cn
}

\begin{document}

\maketitle

\begin{abstract}
In recent years, employing layer attention to enhance interaction among hierarchical layers has proven to be a significant advancement in building network structures. In this paper, we delve into the distinction between layer attention and the general attention mechanism, noting that existing layer attention methods achieve layer interaction on fixed feature maps in a static manner. These static layer attention methods limit the ability for context feature extraction among layers. To restore the dynamic context representation capability of the attention mechanism, we propose a Dynamic Layer Attention (DLA) architecture. The DLA comprises dual paths, where the forward path utilizes an improved recurrent neural network block, named Dynamic Sharing Unit (DSU), for context feature extraction. The backward path updates features using these shared context representations. Finally, the attention mechanism is applied to these dynamically refreshed feature maps among layers. Experimental results demonstrate the effectiveness of the proposed DLA architecture, outperforming other state-of-the-art methods in image recognition and object detection tasks. Additionally, the DSU block has been evaluated as an efficient plugin in the proposed DLA architecture. The code is available at \textcolor{blue}{\url{https://github.com/tunantu/Dynamic-Layer-Attention}}.

\end{abstract}

\section{Introduction}
Numerous studies have highlighted the significance of enhancing interaction among hierarchical layers in Deep Convolutional Neural Networks (DCNNs), which have made substantial progress across various tasks. For instance, ResNet \cite{he2016deep} introduced a straightforward and highly effective implementation by incorporating skip connections between two consecutive layers. DenseNet \cite{huang2017densely} further improved inter-layer interaction by recycling information from all previous layers. Meanwhile, attention mechanisms are playing an increasingly crucial role in DCNNs. The evolution of attention mechanisms in DCNNs has progressed through several stages, including channel attention (\cite{hu2018squeeze}, \cite{wang2020eca}), spatial attention (\cite{carion2020end}, \cite{wang2018non}), branch attention (\cite{srivastava2015training}, \cite{li2019selective}), and temporal attention (\cite{xu2017jointly}, \cite{chen2018video}).

Recently, attention mechanisms have been successfully applied to another direction, (e.g., DIANet \cite{huang2020dianet}, RLANet \cite{zhao2021recurrence}, MRLA \cite{fang2023cross}), indicating the feasibility of strengthening interaction among layers through attention mechanisms. Compared to simple interaction like those in ResNet and DenseNet, the introduction of attention mechanisms makes the interaction among layers more closely effective. DIANet employed a parameter-shared LSTM module along the network's depth to facilitate interaction among layers. RLANet proposed a layer aggregation structure to reuse features of previous layers for enhancing layer interaction. MRLA first introduced the concept of layer attention, treating each feature as a token to learn useful information from others by attention mechanisms.

\begin{figure}
    \centering	
    \begin{minipage}{0.23\textwidth}
    	\includegraphics[width=\textwidth]{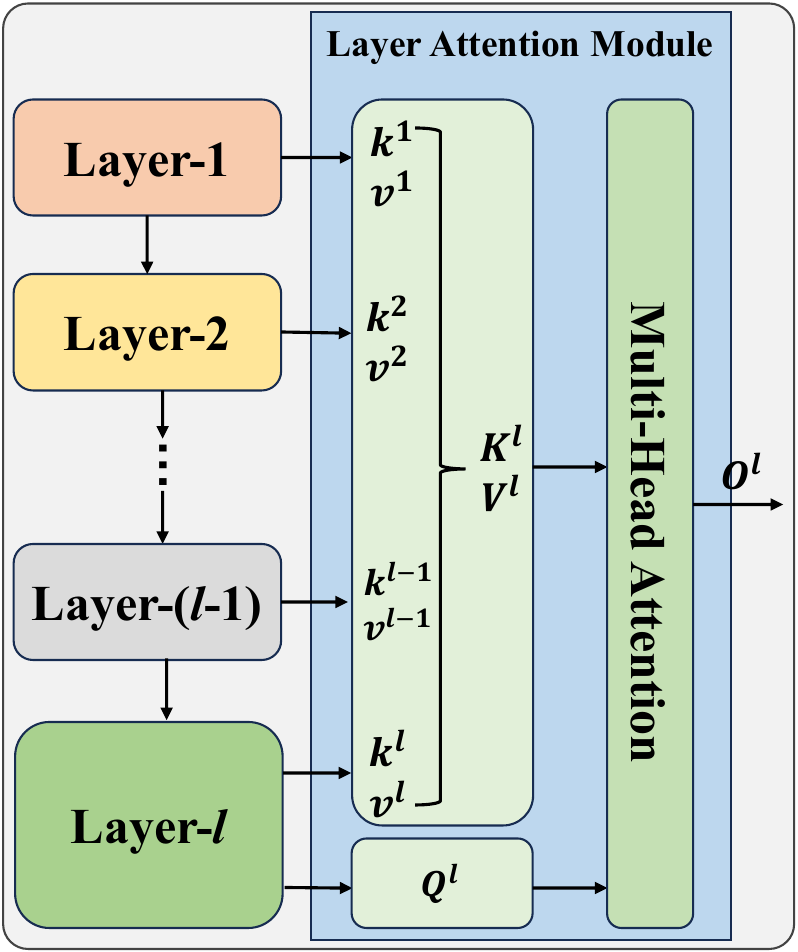}\subcaption{Static Layer Attention}
    \end{minipage}
    \begin{minipage}{0.23\textwidth}
    	\includegraphics[width=\textwidth]{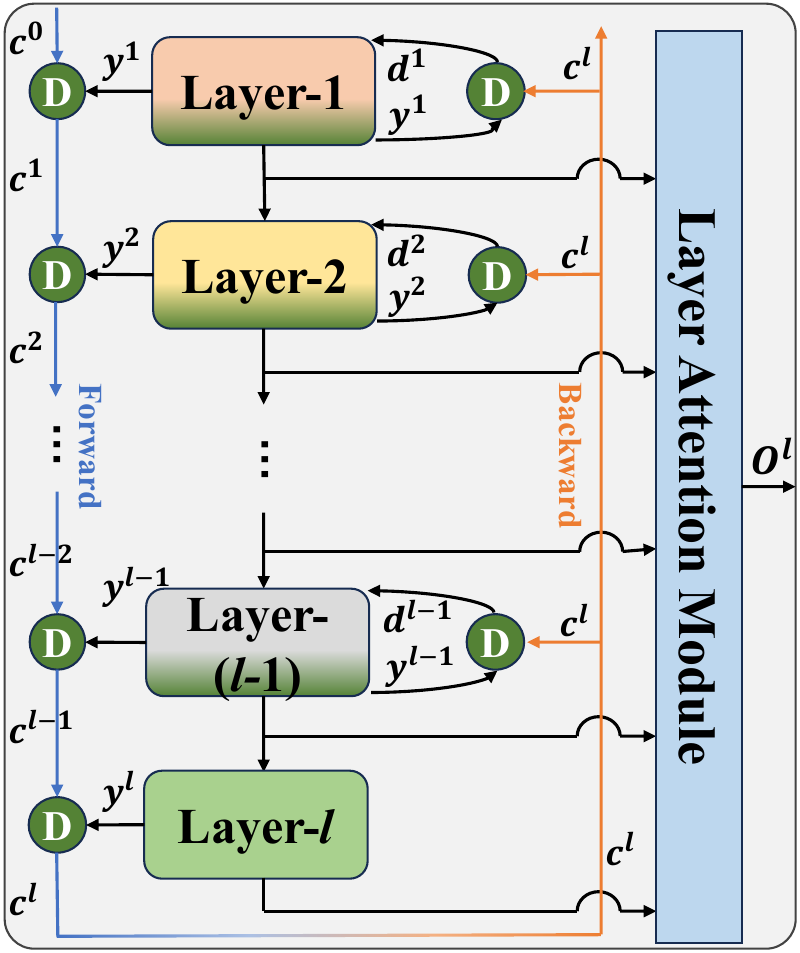}\subcaption{Dynamic Layer Attention}
    \end{minipage}
    \caption{The comparison between \textbf{static} and \textbf{dynamic} layer attention architecture. }\label{fig:1}
\end{figure}

\begin{figure}
    \centering	
    \begin{minipage}{0.23\textwidth}
    	\includegraphics[width=\textwidth]{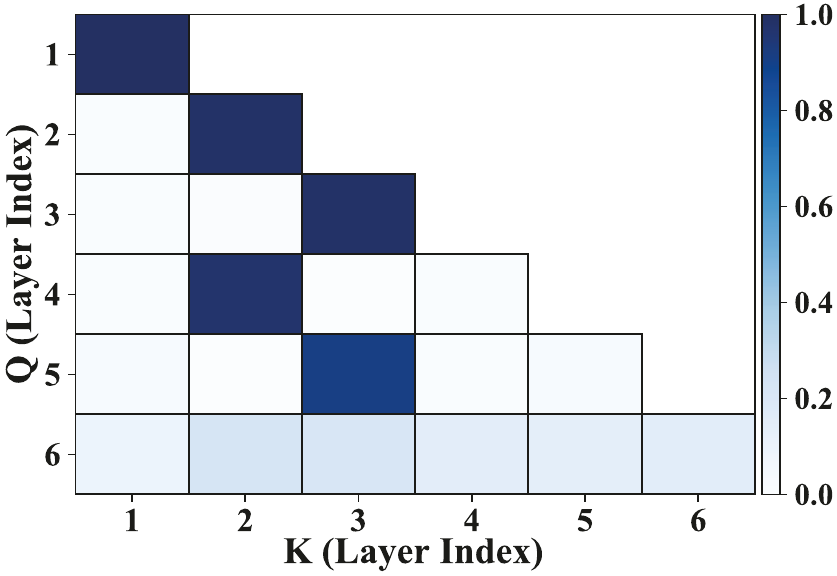}\subcaption{Static Layer Attention}
    \end{minipage}
    \begin{minipage}{0.23\textwidth}
    	\includegraphics[width=\textwidth]{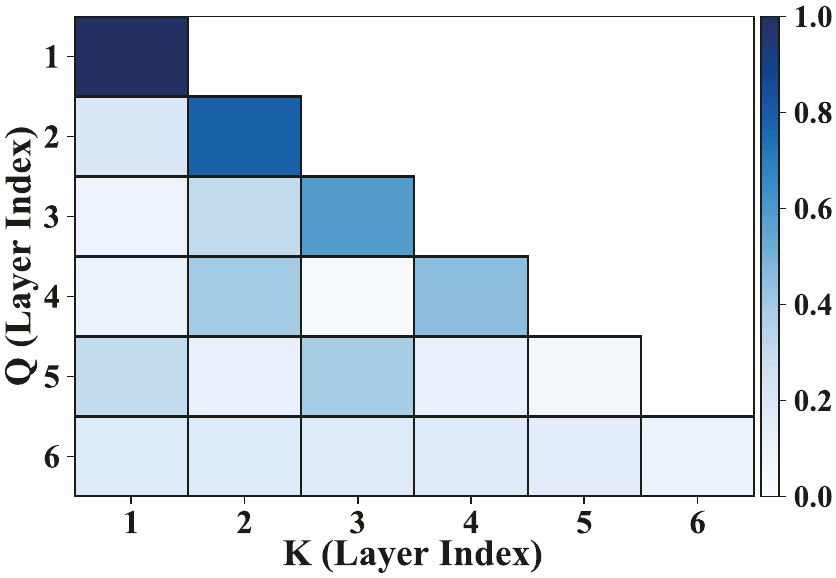}\subcaption{Dynamic Layer Attention}
    \end{minipage}
    \caption{Visualization of attention scores from stage 3 of the ResNet-56 between \textbf{static} and the proposed \textbf{dynamic} layer attention on CIFAR-100 dataset. }\label{fig:2}
\end{figure}

However, we have identified a common drawback in existing layer attention mechanisms: they are applied in a \textbf{static} manner, limiting inter-layer information interaction. In channel and spatial attention, for the input $\boldsymbol{x} \in \mathbb{R}^{C \times H \times W}$, tokens are input to the attention module, all of which are generated from $\boldsymbol{x}$ concurrently. However, in existing layer attention, features generated from different time steps are treated as tokens and passed into the attention module, as shown in Figure \ref{fig:1}(a). Since tokens generated earlier do not change once produced, the input tokens are relatively static, leading to a reduction in information interaction between the current layer and previous layers. Figure \ref{fig:2}(a) visualizes the MRLA attention scores from stage 3 of ResNet-56 trained on CIFAR-100. When the first 5 layers reuse information from previous layers through static layer attention, the key values from only one specific layer are activated, with almost zero attention assigned to other layers. This observation verifies that static layer attention compromises the efficiency of information interaction among layers.

To solve the static problem of layer attention, we propose a novel \textbf{Dynamic Layer Attention} (DLA) architecture to improve information flow among layers, where the information of previous layers can be dynamically modified during the feature interaction. As shown in Figure \ref{fig:2}(b), during the reutilization of information from preceding layers, the attention of the current feature undergo a shift from exclusively focusing on a particular layer to incorporating information from various layers. DLA facilitates a more thorough exploitation of information, enhancing inter-layer information interaction efficiency. Experimental results demonstrate the effectiveness of the proposed DLA architecture, outperforming other state-of-the-art methods in image recognition and object detection tasks. The contributions of this paper are summarized as follows:

\begin{itemize}
    \item We propose a novel DLA architecture, which consists of dual paths, where the forward path extracts context feature among layers using a Recurrent Neural Network (RNN) and the backward path refreshes the original feature at each layer using these shared context representation.
    \item A novel RNN block, named Dynamic Sharing Unit (DSU), is proposed to be a suitable component for DLA. It effectively promotes the dynamic modification of information within DLA and demonstrates commendable performance in the layer-wise information integration as well.
    
\end{itemize}

\section{Related Work}
\textbf{Layer Interaction.}\quad
Layer interaction has consistently been an intriguing aspect of DCNNs. Initially, the implementation of layer interaction was relatively simple. ResNet \cite{he2016deep} introduced skip connections between consecutive layers, mitigating issues of gradient vanishing and exploding to some extent. DenseNet \cite{huang2017densely} further enhanced layer interaction by reusing information generated from all preceding layers. In U-Net \cite{ronneberger2015u}, a commonly utilized architecture in medical segmentation, the encoder and decoder are connected through skip connections to improve feature extraction and achieve higher accuracy.

Recent studies have explored effective methods for implementing layer interaction. DREAL \cite{li2020deep} optimized parameters by introducing arbitrary attention modules and employed a Long Short-Term Memory (LSTM) \cite{hochreiter1997long} to incorporate previous attention weights. The update of parameters for both LSTM and attention layers was achieved through deep reinforcement learning. DIANet \cite{huang2020dianet} incorporated a LSTM module at the layer level, constructing a DIA block shared by all layers in the entire network to facilitate inter-layer interaction. RealFormer \cite{he2020realformer} integrated residual connections between adjacent attention modules, adding the attention scores from the previous layer to the current one. RLANet \cite{zhao2021recurrence} introduced a lightweight recurrent layer aggregation module to describe how information from previous layers can be efficiently reused for better feature extraction in the current layer. \cite{zhao2022ba} proposed a straightforward and versatile approach to strike a balance between effectively utilizing neural network information and maintaining high computational efficiency. By seamlessly integrating features from preceding layers, it foster effective interaction of information. MRLA \cite{fang2023cross} treated the features of each layer as tokens, enabling interaction among different hierarchical layers through an attention mechanism, further strengthening the interaction among layers.

\begin{figure*}[htbp]
    \centering	
    \begin{minipage}{0.31\textwidth}
    	\includegraphics[width=\textwidth]{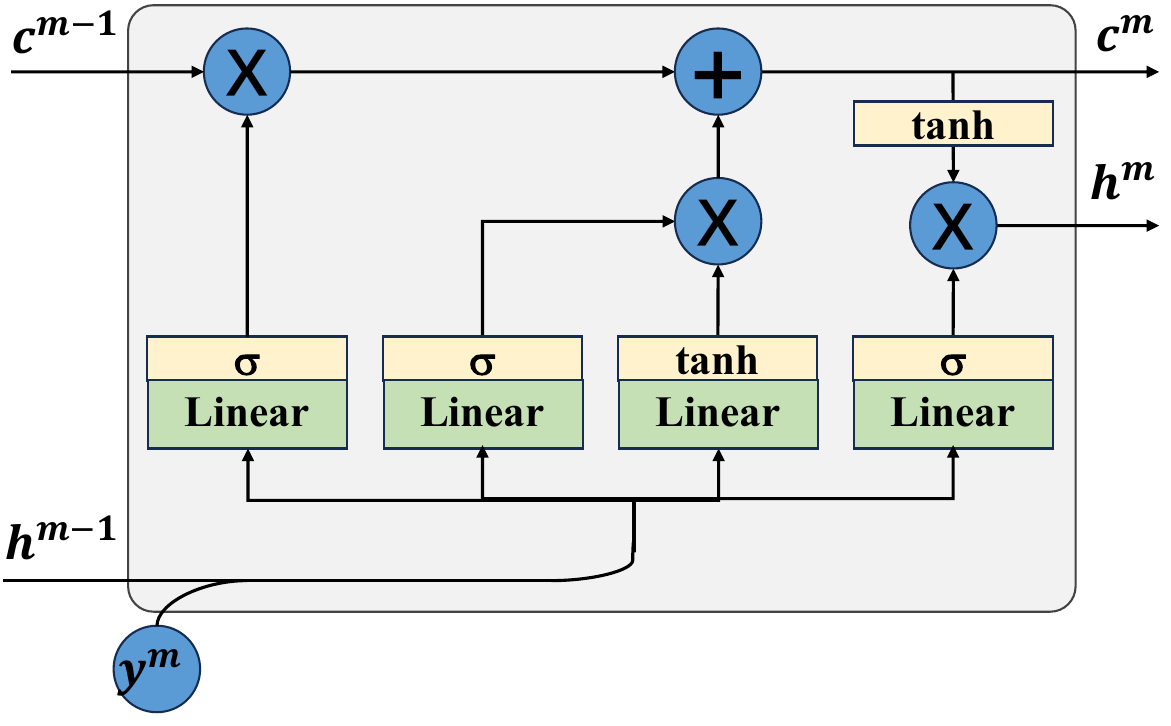}\subcaption{LSTM}
    \end{minipage}\hspace{8pt}
    \begin{minipage}{0.31\textwidth}
    	\includegraphics[width=\textwidth]{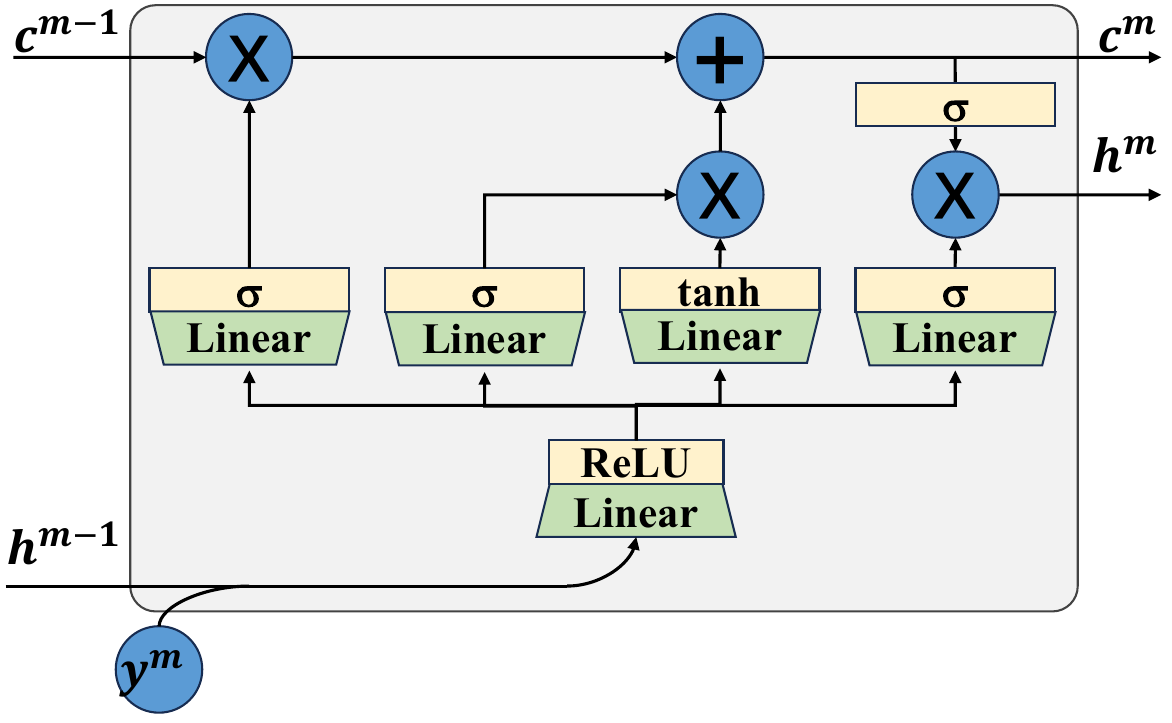}\subcaption{DIA}
    \end{minipage}\hspace{8pt}
    \begin{minipage}{0.31\textwidth}
    	\includegraphics[width=\textwidth]{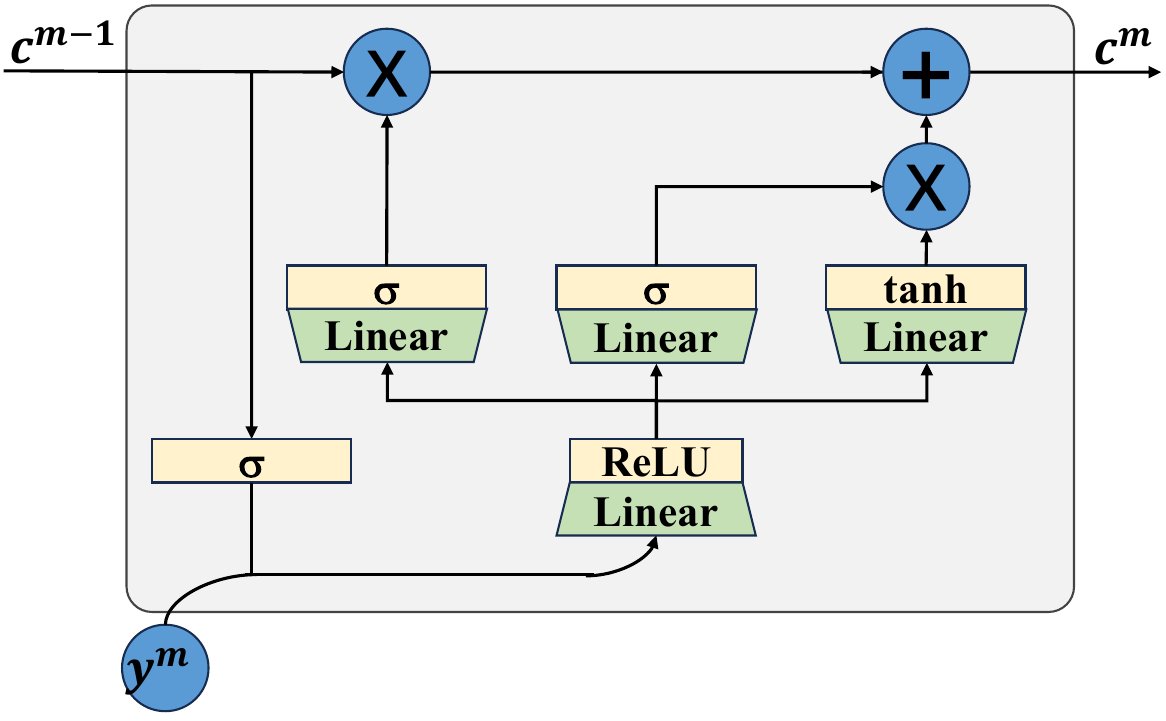}\subcaption{DSU}
    \end{minipage}
    \caption{A comparison of LSTM, DIA, and the proposed DSU blocks. }\label{fig:3}
\end{figure*}

\paragraph{\textbf{Dynamic Network Architecture.}}
Dynamic networks represent a type of neural network structure whose topology or parameters can dynamically change during runtime, providing the network with considerable flexibility or other benefits. \cite{wang2020glance} proposed a universally adaptable inference framework for the majority of DCNNs, with costs that can be easily dynamically adjusted without additional training. RANet \cite{yang2020resolution} introduced an adaptive network that could effectively reduce the spatial redundancy involved in inferring high-resolution inputs. \cite{han2021dynamic} have shown that dynamic neural networks can enhance the efficiency of deep networks. CondenseNetv2 \cite{yang2021condensenet} utilized a Sparse Feature Reactivation (SFR) module to reactivate pruned features from CondenseNet \cite{huang2018condensenet}, thereby enhancing feature utilization efficiency. To the best of our knowledge, this paper is the first attempt to build a dynamic network architecture for strengthening layer interaction.

\section{Dynamic Layer Attention}
We will start by reconsidering the current layer attention architecture and elucidating its static nature. Subsequently, we will introduce the Dynamic Layer Attention (DLA). Finally, we will present an enhanced RNN plugin block named the Dynamic Sharing Unit (DSU), integrated within the DLA architecture.

\subsection{Rethinking Layer Attention}
Layer attention was recently defined by \cite{fang2023cross} and is illustrated in Figure \ref{fig:1}(a), where the attention mechanism enhances layer interaction. \cite{fang2023cross} focused on reducing the computational cost of layer attention and proposed the Recurrent Layer Attention (RLA) architecture. In RLA, features from distinct layers are treated as tokens and undergo computations, ultimately producing attention output. Let the feature output of the $l$-th layer be $\boldsymbol{x}^l \in \mathbb{R}^{C \times W \times H}$. The vectors $\boldsymbol{Q}^l$, $\boldsymbol{K}^l$, and $\boldsymbol{V}^l$ can be calculated as follows:
\begin{equation}\label{eqn:qkv}
\left\{
\begin{aligned}
\boldsymbol{Q}^l &= f^l_q(\boldsymbol{x}^l)\\
\boldsymbol{K}^l &= \text{Concat}\left[f^1_k(\boldsymbol{x}^1), \ldots, f^l_k(\boldsymbol{x}^l)\right] \\
\boldsymbol{V}^l &= \text{Concat}\left[f^1_v(\boldsymbol{x}^1), \ldots, f^l_v(\boldsymbol{x}^l)\right],
\end{aligned}
\right.
\end{equation}
where $f_q$ is a mapping function to extract information from the $l$-th layer, while $f_k$ and $f_v$ are corresponding mapping functions intended to extract information from the $1$st to $l$-th layers, respectively. The attention output $\boldsymbol{o}^l$ is given as follows:
\begin{equation}\label{eqn:base}
\begin{aligned}
    \boldsymbol{o}^l &= \text{Softmax}\left(\frac{\boldsymbol{Q}^l (\boldsymbol{K}^l)^\text{T}}{\sqrt{D_k}}\right) \boldsymbol{V}^l \\
    &=\sum^l_{i=1} \text{Softmax}\left(\frac{\boldsymbol{Q}^l \left[f_k^i(\boldsymbol{x}^i)\right]^\text{T}}{\sqrt{D_k}}\right) f_v^i(\boldsymbol{x}^i),
\end{aligned}
\end{equation}
where \(D_k\) serves as the scaling factor. A lightweight version of RLA was proposed to recurrently update the attention output $\boldsymbol{o}^l$ as follows:
\begin{equation}\label{eqn:light}
    \boldsymbol{o}^l = \boldsymbol{\lambda}_o^l \odot  \boldsymbol{o}^{l-1} + \text{Softmax}\left(\frac{\boldsymbol{Q}^l \left[f_k^l(\boldsymbol{x}^l)\right]^\text{T}}{\sqrt{D_k}}\right) f_v^l(\boldsymbol{x}^l),
\end{equation}
where $\boldsymbol{\lambda}^{l}_o$ is a learnable vector and $\odot$ indicates the element-wise multiplication. With the multi-head structure design, Multi-head RLA (MRLA) is introduced.

\subsection{Motivation} 
MRLA successfully integrated the attention mechanism to enhance layer interaction, effectively addressing computational costs. However, when MRLA is applied in the $l$-th layer, a preceding feature output \(\boldsymbol{x}^m\) (\(m<l\)) has already been generated in the $m$-th layer, with no subsequent changes. Consequently, the information processed by MRLA comprises fixed features from the previous layers. In contrast, widely used attention-based models, such as channel attention \cite{hu2018squeeze,wang2020eca}, spatial attention \cite{carion2020end,huang2019ccnet}, and Transformers \cite{vaswani2017attention,liu2021swin,chen2023dual,jiao2023dilateformer}, pass tokens into the attention module generated simultaneously. Applying the attention module between freshly generated tokens ensures that each token consistently learns up-to-date features. Therefore, we categorize MRLA as a \textbf{static} layer attention mechanism, limiting interaction between the current layer and shallower layers.

% \begin{figure*}[htbp]
%     \centering	
%     \begin{minipage}{0.31\textwidth}
%     	\includegraphics[width=\textwidth]{LSTM.pdf}\subcaption{LSTM}
%     \end{minipage}\hspace{8pt}
%     \begin{minipage}{0.31\textwidth}
%     	\includegraphics[width=\textwidth]{DIA.pdf}\subcaption{DIA}
%     \end{minipage}\hspace{8pt}
%     \begin{minipage}{0.31\textwidth}
%     	\includegraphics[width=\textwidth]{DSU.pdf}\subcaption{DSU}
%     \end{minipage}
%     \caption{A comparison of LSTM, DIA, and the proposed DSU blocks. }\label{fig:2}
% \end{figure*}

In a general self-attention mechanism, the feature
\(\boldsymbol{x}^m\) serves two purposes: conveying essential information and representing context. The essential information extracted at the current layer distinguishes it from that at other layers. Meanwhile, the context representation captures changes and evolution of features along the temporal axis, a critical aspect in determining feature freshness. In the general attention mechanism, essential information is generated at each layer, and the context representation is transferred to the next layer for calculating the attention output. In contrast, in layer attention, once tokens are generated, attention is calculated with a fixed context representation, diminishing the efficiency of the attention mechanism. Therefore, this paper aims to establish a novel method to restore the context representation, ensuring that the information fed into the layer attention is consistently dynamic.

\subsection{Dynamic Layer Attention Architecture}
To address the static issue of MRLA, we propose the use of a dynamic updating rule to extract the context representation and promptly update features at previous layers, resulting in a Dynamic Layer Attention (DLA) architecture. As illustrated in Figure \ref{fig:1}(b), DLA consists of dual paths: forward and backward. In the forward path, a Recurrent Neural Network (RNN) is employed for context feature extraction. Let the RNN block be denoted as ``Dyn", and the initial context as $\boldsymbol{c}^0$, respectively. $\boldsymbol{c}^0$ is randomly initialized. Given an input $\boldsymbol{x}^m\in \mathbb{R}^{
C\times W\times H}$ where $m< l$, a Global Average Pooling (GAP) is applied to extract global features at $m$-th layer as follows:
\begin{equation}
    \boldsymbol{y}^m = \text{GAP}(\boldsymbol{x}^m),\ \boldsymbol{y}^m \in \mathbb{R}^{C}. 
\end{equation}

The context representation is extracted as follows:
\begin{equation}\label{eqn:dyn}
\boldsymbol{c}^m = \text{Dyn}(\boldsymbol{y}^m, \boldsymbol{c}^{m-1}; \theta^l).
\end{equation}
where $\theta^l$ represents the shared trainable parameters of ``Dyn". Once the context $\boldsymbol{c}^l$ is calculated, the features of each layer are simultaneously updated in the backward path as follows:
\begin{equation}\label{eqn:update}
    \left\{
    \begin{aligned}
        \boldsymbol{d}^m &= \text{Dyn}(\boldsymbol{y}^m, \boldsymbol{c}^l; \theta^l)\\
        \boldsymbol{x}^m &\leftarrow \boldsymbol{x}^m \odot \boldsymbol{d}^m
    \end{aligned}\right.
\end{equation}

Referring to Equation (\ref{eqn:dyn}), the forward context feature extraction is a step-by-step process with a computation complexity of $\mathcal{O}(n)$. Meanwhile, the feature updating in Equation (\ref{eqn:update}) can be performed in parallel, resulting in a computation complexity of $\mathcal{O}(1)$. After updating $\boldsymbol{x}^m$, the basic version of DLA use Equation (\ref{eqn:base}) to compute the layer attention, abbreviated as DLA-B. For the lightweight version of DLA, Simply update $\boldsymbol{o}^{l-1}$ and then use Equation (\ref{eqn:light}) to obtain DLA-L.

% After updating $\boldsymbol{x}^m$, the basic and lightweight versions of DLA use Equation (\ref{eqn:base}) and (\ref{eqn:light}) to compute the layer attention, abbreviated as DLA-B and DLA-L, respectively.

\paragraph{\textbf{Computation Efficiency.}}
DLA possesses several advantages in its structural design. Firstly, global information is condensed to compute context information, a utility that has been validated in \cite{huang2020dianet}. Secondly, DLA employs shared parameters within a RNN block. Thirdly, the context $\boldsymbol{c}^l$ is separately fed into the feature maps at each layer in parallel. Both the forward and backward paths share the same parameters throughout the entire network. Finally, we introduce an efficient RNN block for calculating context representation, which will be elucidated in the following subsection. With these efficiently designed structural rules, the computation cost and network capacity are guaranteed.
  
\subsection{Dynamic Sharing Unit}
LSTM, as depicted in Figure \ref{fig:2}(a), is designed for processing sequential data and learning temporal features, enabling it to capture and store information over long sequences. However, the fully connected linear transformation in LSTM significantly increases the network capacity when embedding LSTM as the recurrent block in DLA. To mitigate this capacity increase, a variant LSTM block named the DIA unit was proposed by \cite{huang2020dianet}, as illustrated in Figure \ref{fig:2}(b). Before feeding data into the network, DIA first utilizes a linear transformation followed by a ReLU activation function to reduce the input dimension. Additionally, DIA replaces the Tanh function with a Sigmoid function at the output layer.

LSTM and DIA generate two outputs, comprising a hidden vector $\boldsymbol{h}^m$ and a cell state vector $\boldsymbol{c}^m$. Typically, $\boldsymbol{h}^m$ is used as the output vector, and $\boldsymbol{c}^m$ serves as the memory vector. DLA is exclusively focused on extracting context characteristics from different layers, where the RNN block has no duty to transform its internal state feature to the outside. Consequently, we discard the output gate and merge the memory and hidden vector by omitting the $\boldsymbol{h}^m$ symbol. The proposed simplified RNN block is named Dynamic Sharing Unit (DSU). The workflow of the DSU is illustrated in Figure \ref{fig:2}(c). Specifically, before adding $\boldsymbol{c}^{m-1}$ and $\boldsymbol{y}^m$, we first normalize $\boldsymbol{c}^{m-1}$ using an activation function $\sigma(\cdot)$. Here, we opt for the Sigmoid function (\(\sigma(z) = 1
/(1 + e^{-z})\)). Therefore, the input to DSU was compressed as follows:
\begin{equation}
    \boldsymbol{s}^m = \text{ReLU}\left(\boldsymbol{W}_1\left[ \sigma(\boldsymbol{c}^{m-1}), \boldsymbol{y}^m \right] \right).
\end{equation}

The hidden transformation, the input gate, and the forget gate can be represented by the following formula:
\begin{equation}
    \left\{
    \begin{aligned}
    \boldsymbol{\tilde{c}}^m &= \text{Tanh}(\boldsymbol{W}_2^c \cdot \boldsymbol{s}^m + b^c) \\
    \boldsymbol{i}^m &= \sigma(\boldsymbol{W}_2^i \cdot \boldsymbol{s}^m + b^i ) \\
    \boldsymbol{f}^m &= \sigma(\boldsymbol{W}_2^f \cdot \boldsymbol{s}^m + b^f )
    \end{aligned}
    \right.
\end{equation}

Subsequently, we obtain 
\begin{equation}
\boldsymbol{c}^m = \boldsymbol{f}^m \odot \boldsymbol{c}^{m-1} + \boldsymbol{i}^m \odot \boldsymbol{\tilde{c}}^m
\end{equation}

To decrease the network parameters, let $\boldsymbol{W}_1\in \mathbb{R}^{\frac{C}{r}\times 2C}$ and $\boldsymbol{W}_2\in \mathbb{R}^{C\times \frac{C}{r}}$, where $r$ is the reduction ratio. DSU reduces the parameters to $5C^2/r$, which is fewer than $8C^2$ of LSTM and $10C^2/r$ of DIA.

\section{Experiments}
\subsection{Image Classification}

\textbf{Experimental Setup.}\quad
We conducted experiments on the CIFAR-10, CIFAR-100, and ImageNet-1K datasets using ResNets \cite{he2016deep} as the backbone network for image classification. For the CIFAR-10 and CIFAR-100 datasets, we employed standard data augmentation strategies \cite{huang2016deep}. The training process involved random horizontal flipping of images, padding each side by 4 pixels, and then randomly cropping to 32$\times$32. Normalization with mean and standard deviation adjustment was implemented, and training hyperparameters such as batch size, initial learning rate, and weight decay followed the recommendations of the original ResNets \cite{he2016deep}. To address hyperparameter uncertainty, we conducted five runs of experiments. For the ImageNet-1K dataset, we adopted the same data augmentation strategy and hyperparameter settings outlined in \cite{he2016deep} and \cite{he2019bag}. During training, images were randomly cropped to 224$\times$224 with horizontal flipping. In the testing phase, images were resized to 256×256, then centrally cropped to a final size of 224$\times$224. The optimization process utilized an SGD optimizer with a momentum of 0.9 and weight decay of 1e-4. The initial learning rate was set to 0.1 and decreased according to the MultiStepLR schedule over 100 epochs for a batch size of 256, consistent with the ResNet approach \cite{he2016deep}. Meanwhile, the reduction ratio $r$ was set to 4 for the CIFAR-10 and CIFAR-100 datasets, and 20 for the ImageNet-1K dataset, consistent with the settings in DIANet \cite{huang2020dianet}.

\begin{figure}[htbp]
    \centering	
    \begin{minipage}{0.235\textwidth}
    	\includegraphics[width=\textwidth]{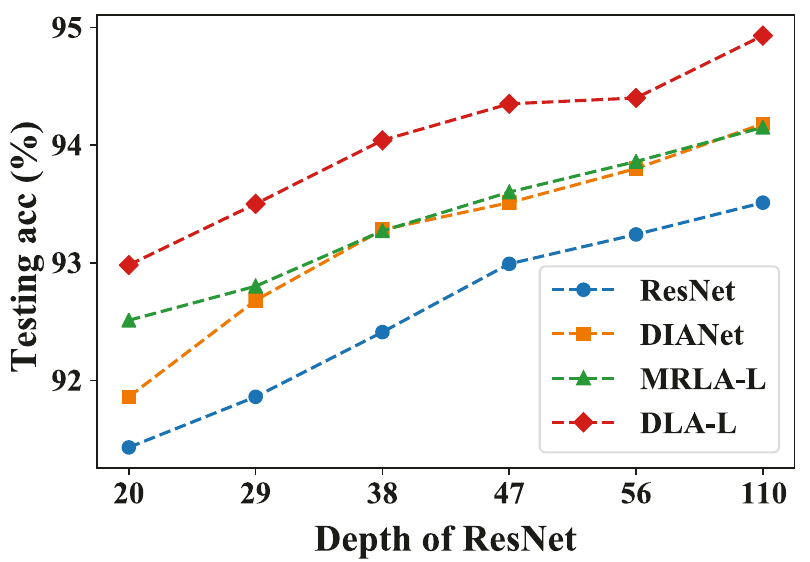}\subcaption{Testing acc. on CIFAR-10}
    \end{minipage}
    \begin{minipage}{0.235\textwidth}
        \includegraphics[width=\textwidth]{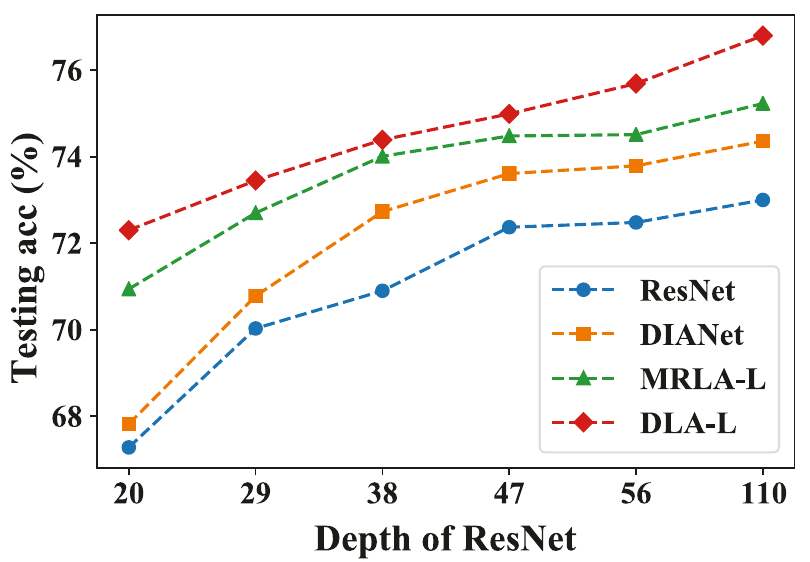}\subcaption{Testing acc. on CIFAR-100}
    \end{minipage}
    \caption{Comparison of testing accuracy of different models with different depths. }\label{fig:4}
\end{figure}

\paragraph{\textbf{Results on CIFAR.}}
 The experimental results, showcasing Accuracy$\pm$Std, are presented in Table \ref{tab:1}. These results underscore the significant superiority of our DLA-B and DLA-L models over other challenging networks, including SE \cite{hu2018squeeze}, ECA \cite{wang2020eca}, DIANet \cite{huang2020dianet}, MRLA \cite{fang2023cross} on the CIFAR-10 and CIFAR-100 datasets. In comparison with the baselines, DLA-L's top-1 accuracy on CIFAR-10 surpasses ResNets by 1.32\%, 1.60\%, and 1.62\%, and even outperforms them on CIFAR-100 by 4.96\%, 2.94\%, and 3.41\%. Furthermore, both DLA-B and DLA-L outperform MRLA-B and MRLA-L, which are typical static layer attention models. It is noteworthy that DLA-L-20 (embedding ResNet-20) achieves fewer parameters than ResNet-56 while maintaining comparable top-1 accuracy on the CIFAR-10 (92.96\% vs. 92.95\%) and CIFAR-100 (72.26\% vs. 72.32\%) datasets. Furthermore, DLA-L-56 outperforms ResNet-110 by 1.51\% and 2.46\% on CIFAR-10 and CIFAR-100 datasets, respectively. As depicted in Figure \ref{fig:4}, ResNet, DIANet, and MRLA-L exhibit a relatively slow growth in capabilities with increasing network depth. In contrast, the proposed DLA demonstrates a faster increase in test accuracy on the CIFAR-10 and CIFAR-100 datasets as the network depth increases. This observation verifies that strengthening layer interaction through DLA is more beneficial in a deeper network structure.

 \begin{table}[htbp]
\centering
\setlength\tabcolsep{3pt} 
\begin{tabular}{l|cc|cc}
\hline
\multicolumn{1}{c|}{\multirow{2}{*}{\textbf{\diagbox{Model}{Data}}}} & \multicolumn{2}{c|}{\textbf{CIFAR-10}} & \multicolumn{2}{c}{\textbf{CIFAR-100}} \\
\cline{2-5}
\multicolumn{1}{c|}{} & \textbf{\#P(M)} & \textbf{Top-1 } & \textbf{\#P(M)} & \textbf{Top-1 } \\
\hline
\textbf{ResNet-20} & 0.22 & 91.64$\pm$0.18 & 0.24 & 67.30$\pm$0.28 \\
SE & 0.24 & 91.29$\pm$0.24 & 0.27 & 68.93$\pm$0.35 \\
ECA & 0.22 & 91.63$\pm$0.16 & 0.24 & 67.23$\pm$0.24 \\
DIANet & 0.44 & 91.43$\pm$0.14 & 0.46 & 67.67$\pm$0.22 \\
MRLA-B & 0.23 & 92.15$\pm$0.23 & 0.25 & 71.44$\pm$0.49 \\
DLA-B (ours)& 0.41 & \textbf{92.47$\pm$0.10} & 0.43 & \textbf{72.01$\pm$0.37} \\
MRLA-L & 0.23 & 92.65$\pm$0.08 & 0.25 & 71.46$\pm$0.27 \\
DLA-L (ours)& 0.41 & \textbf{92.96$\pm$0.18} & 0.43 & \textbf{72.26$\pm$0.29} \\
\hline
\textbf{ResNet-56} & 0.59 & 92.95$\pm$0.18 & 0.61 & 72.32$\pm$0.36 \\
SE & 0.66 & 93.60$\pm$0.18 & 0.68 & 73.51$\pm$0.28 \\
ECA & 0.59 & 93.72$\pm$0.14 & 0.61 & 72.63$\pm$0.35 \\
DIANet & 0.81 & 93.88$\pm$0.21 & 0.83 & 73.87$\pm$0.27 \\
MRLA-L & 0.62 & 94.28$\pm$0.26 & 0.65 & 74.18$\pm$0.17 \\
DLA-L (ours) & 0.80 & \textbf{94.55$\pm$0.13} & 0.82 & \textbf{75.46$\pm$0.26} \\
\hline
\textbf{ResNet-110} & 1.15 & 93.04$\pm$0.33 & 1.17 & 73.00$\pm$0.36 \\
SE & 1.28 & 94.09$\pm$0.11 & 1.30 & 75.01$\pm$0.20 \\
ECA & 1.15 & 93.76$\pm$0.31 & 1.17 & 73.97$\pm$0.36 \\
DIANet & 1.37 & 94.48$\pm$0.08 & 1.39 & 75.31$\pm$0.16 \\
MRLA-L & 1.21 & 94.49$\pm$0.31 & 1.24 & 75.16$\pm$0.24 \\
DLA-L (ours)& 1.39 & \textbf{94.66$\pm$0.23} & 1.41 & \textbf{76.41$\pm$0.36} \\
\hline
\end{tabular}
\caption{Testing accuracy (\%) on CIFAR-10 and CIFAR-100 datasets. ``\#P(M)'' means the number of parameters (million).}\label{tab:1}
\end{table}

\begin{table}[htbp]
  \centering
  \setlength\tabcolsep{2.5mm}{ 
  \begin{tabular}{lcccc}
    \toprule
    \textbf{Model} & \textbf{Params} & \textbf{FLOPs} & \textbf{Top-1} & \textbf{Top-5} \\
    \midrule
    \textbf{ResNet-50}  & 25.6M & 4.1B & 76.1 & 92.9 \\
    SE  & 28.1M & 4.1B & 76.7 & 93.4 \\
    CBAM  & 28.1M & 4.2B & 77.3 & 93.7 \\
    $A^2$  & 34.6M & 7.0B & 77.0 & 93.5 \\
    AA  & 27.1M & 4.5B & 77.7 & 93.8 \\
    all GC  & 29.4M & 4.2B & 77.7 & 93.7 \\
    ECA & 25.6M & 4.1B & 77.5 & 93.7 \\
    DIANet  & 28.4M & - & 77.2 & - \\
    RLA$_g$ & 25.9M & 4.5B & 77.2 & 93.4 \\
    MRLA-L  & 25.7M & 4.2B & 77.7 & 93.8 \\
    DLA-L (ours) & 27.2M & 4.3B & \textbf{78.0} & \textbf{94.0} \\ 
    \midrule
    \textbf{ResNet-101}  & 44.5M & 7.8B & 77.4 & 93.5 \\
    SE  & 49.3M & 7.8B & 77.6 & 93.9 \\
    CBAM  & 49.3M & 7.9B & 78.5 & 94.3 \\
    AA  & 47.6M & 8.6B & 78.7 & 94.4 \\
    ECA  & 44.5M & 7.8B & 78.7 & 94.3 \\
    RLA$_g$  & 45.0M & 8.4B & 78.5 & 94.2 \\
    MRLA-L  & 44.9M & 7.9B & 78.7 & 94.4 \\
    DLA-L (ours) & 47.8M & 8.1B & \textbf{78.9} & \textbf{94.5} \\
    \bottomrule
  \end{tabular}}
  \caption{Comparisons of accuracy (\%) on the ImageNet-1K validation set (All results of the following methods are captured from their original papers).}\label{tab:2}
\end{table}

\paragraph{\textbf{Results on ImageNet-1K.}}
We compare the DLA-L architecture with other challenging methods using ResNets as baselines. Experimental results, shown in Table \ref{tab:2}, indicate that our model significantly outperforms other models. Firstly, our DLA-L exhibits an increase of 1.9\% and 1.5\% in top-1 accuracy to ResNet-50 and ResNet-101, respectively. Then, when compared to the channel attention method, DLA-L-50 has 0.9M fewer parameters than SE-50 and CBAM-50 \cite{woo2018cbam}, but its top-1 accuracy is higher by 1.3\% and 0.7\%. DLA-L-101 has 1.5M fewer parameters than SE-101 and CBAM-101 \cite{woo2018cbam} while achieving a 1.3\% and 0.4\% top-1 accuracy increase, respectively. Meanwhile, when compared to the lightweight model, DLA-L-50 and DLA-L-101 remains a 0.5\% and 0.2\% higher top-1 accuracy than ECA-50 and ECA-101. DLA-L also outperforms other popular layer interaction models. With 1.2M fewer parameters than DIANet-50 , DLA-L-50 achieves a 0.8\% higher top-1 accuracy than DIANet-50. Although DLA-L introduces more parameters compared to RLA$_g$ \cite{zhao2021recurrence}, it achieves a 0.8\% and 0.4\% higher top-1 accuracy than RLA$_g$-50 and RLA$_g$-101, respectively. Additionally, DLA-L also surpasses recent models such as $A^2$ \cite{chen20182}, AA \cite{bello2019attention}, and GC \cite{cao2019gcnet}. Finally, in comparison with static layer attention, DLA-L exhibits an increase of 0.3\% and 0.2\% in top-1 accuracy to MRLA-L-50 and MRLA-L-101 with a tolerable increase in the number of parameters. In summary, through comparisons with well-known channel attention models, layer interaction models, and various other challenging models, we have validated that our proposed DLA architecture serves as a more effective layer interaction model, outperforming other models in the domain of image classification.

\begin{table*}[ht]
  \centering
    \setlength{\tabcolsep}{3.5mm}{
    \begin{tabular}{l|c|c|ccc|ccc}
    \hline
    
    \hline
    \multicolumn{1}{c|}{\textbf{Methods}} & \textbf{Detectors} & \textbf{Params} & \textbf{AP} & $\textbf{AP}_{50}$ & $\textbf{AP}_{75}$ & $\textbf{AP}_{S}$ & $\textbf{AP}_{M}$ & $\textbf{AP}_{L}$ \bigstrut\\
    \hline
    
    \hline
    \textbf{ResNet-50} \cite{he2016deep} & \multicolumn{1}{c|}{\multirow{12}[3]{*}{\makecell[c]{Faster \\ R-CNN} }} & 41.5M & 36.4  & 58.2  & 39.2  & 21.8  & 40.0    & 46.2 \bigstrut[t]\\
    SE \cite{hu2018squeeze}    &       & 44.0M & 37.7  & 60.1  & 40.9  & 22.9  & 41.9  & 48.2 \\
    ECA \cite{wang2020eca}   &       & 41.5M & 38.0    & 60.6  & 40.9  & 23.4  & 42.1  & 48.0 \\
    RLA$_g$ \cite{zhao2021recurrence}  &       & 41.8M & 38.8  & 59.6  & 42.0    & 22.5  & 42.9  & 49.5 \\
    BA \cite{zhao2022ba}  &       & 44.7M & 39.5  & 61.3  & 43.0 & 24.5 & 43.2 & 50.6 \\
    MRLA-L \cite{fang2023cross} &       & 41.7M & 40.4  & 61.5  & 44.0    & 24.2  & 44.1  & 52.7 \\
    DLA-L (ours) &    & 44.2M  &   \textbf{40.6}    &   \textbf{61.6}    &  \textbf{44.2}     &   \textbf{24.5}    &   \textbf{44.2}    &   \textbf{52.9}     \bigstrut[b]\\
\cline{1-1}\cline{3-9}    \textbf{ResNet-101} \cite{he2016deep} &       & 60.5M & 38.7  & 60.6  & 41.9  & 22.7  & 43.2  & 50.4 \bigstrut[t]\\
    SE \cite{hu2018squeeze}   &       & 65.2M & 39.6  & 62.0    & 43.1  & 23.7  & 44.0    & 51.4 \\
    ECA \cite{wang2020eca}   &       & 60.5M & 40.3  & 62.9  & 44.0    & 24.5  & 44.7  & 51.3 \\
    RLA$_g$ \cite{zhao2021recurrence}  &       & 60.9M & 41.2  & 61.8  & 44.9  & 23.7  & 45.7  & 53.8 \\
    MRLA-L \cite{fang2023cross} &       & 60.9M & 42.0    & 63.1  & 45.7  & 25.0    & 45.8  & 55.4 \\
    DLA-L (ours) &       &   63.4M   &  \textbf{42.3}  &  \textbf{63.3} &  \textbf{45.8}  & \textbf{25.2} &  \textbf{46.0} & \textbf{55.5}  \bigstrut[b]\\
    \hline
    
    \hline
   \textbf{ResNet-50} \cite{he2016deep} & \multicolumn{1}{c|}{\multirow{16}[3]{*}{\makecell[c]{Mask \\ R-CNN}}} & 44.2M & 37.2  & 58.9  & 40.3  & 34.1  & 55.5    & 36.2 \bigstrut[t]\\
    SE \cite{hu2018squeeze}    &       & 46.7M & 38.7  & 60.9 &42.1 & 35.4  & 57.4  & 37.8    \\
    ECA \cite{wang2020eca}  &       & 44.2M & 39.0    & 61.3  & 42.1  & 35.6  & 58.1  & 37.7 \\
    NL \cite{wang2018non}    &        &  46.5M     &38.0  & 59.8   & 41.0 & 34.7 &  56.7 & 36.6  \\
    GC \cite{cao2019gcnet}   &        &   54.4M     &  39.9 & 62.2 & 42.9 & 36.2  & 58.7 &  38.3 \\
    RLA$_g$ \cite{zhao2021recurrence}   &   & 44.4M & 39.5  & 60.1  & 43.4 & 35.6  & 56.9  & 38.0 \\
    BA \cite{zhao2022ba}  &        &    47.3M    &  40.5 &  61.7 &  44.2 &  36.6  & 58.7 &  38.6 \\
    MRLA-L \cite{fang2023cross} &       & 44.3M & 41.2  & 62.3 & 45.1  & 37.1  & 59.1 & 39.6 \\
    DLA-L (ours) &       &  46.9M & \textbf{41.4}  &   \textbf{62.5}    &   \textbf{45.3}    &   \textbf{37.2}    &  \textbf{59.3}   &  \textbf{39.7}      \bigstrut[b]\\
\cline{1-1}\cline{3-9}    \textbf{ResNet-101} \cite{he2016deep} &       & 63.2M & 39.4  & 60.9  & 43.3  & 35.9  & 57.7  & 38.4 \bigstrut[t]\\
    SE \cite{hu2018squeeze}   &       & 67.9M & 40.7  & 62.5 & 44.3  & 36.8  & 59.3  & 39.2 \\
    ECA \cite{wang2020eca}  &       & 63.2M & 41.3 & 63.1  & 44.8  & 37.4  & 59.9  & 39.8 \\
    NL \cite{wang2018non}   &        &  65.5M & 40.8 &  63.1  & 44.5   & 37.1 & 59.9 & 39.2  \\
    GC \cite{cao2019gcnet}   &        &  82.2M  & 41.7 & 63.7 &  45.5 &  37.6  & 60.5 &  39.8 \\
    RLA$_g$ \cite{zhao2021recurrence}   &       & 63.6M & 41.8  & 62.3  & 46.2   & 37.3  & 59.2 & 40.1 \\
    MRLA-L \cite{fang2023cross} &       & 63.5M & 42.8  & 63.6  & 46.5   & 38.4  & 60.6  & 41.0 \\
    DLA-L (ours) &       &   66.1M   &   \textbf{42.9}    &     \textbf{63.8}  & \textbf{46.7}  &     \textbf{38.6}  & \textbf{60.9}  & \textbf{41.2} \bigstrut[b]\\
    \hline
    
    \hline
    \end{tabular}}
    \caption{The object detection results on the COCO val2017 with Faster R-CNN and Mask R-CNN.}
  \label{tab:3}
\end{table*}

\subsection{Object Detection}
\textbf{Experimental Setup.}\quad 
In the context of object detection, our approach was evaluated on the COCO2017 dataset using the Faster R-CNN \cite{ren2015faster} and Mask R-CNN \cite{he2017mask} frameworks as detectors. The implementations of all detectors were carried out using the MMDetection toolkit \cite{chen2019mmdetection}, following the default settings. In the preprocessing stage, the shorter side of input images was resized to 800 pixels. The optimization process employed SGD with a weight decay of 1e-4, momentum of 0.9, and a batch size of 8. The models underwent training for a total of 12 epochs, starting with an initial learning rate of 0.01. Learning rate adjustments occurred at the 8th and 11th epochs, with a reduction by a factor of 10 each time.

\paragraph{\textbf{Results on COCO2017.}}
As shown in Table \ref{tab:3}, when utilizing Faster R-CNN as the detectors, our proposed DLA-L demonstrates a remarkable improvement in average precision (AP) of 4.2\% and 3.6\% on ResNet-50 \cite{he2016deep} and ResNet-101 \cite{he2016deep}, respectively, which validates the outstanding capability of the DLA architecture in enhancing object detection performance. Notably, DLA-L outperforms other models, when compared to the challenging channel attention block, DLA-L-50 exhibits similar parameter counts to SE-50 \cite{hu2018squeeze} but achieves a higher AP by 2.9\% and a 3.3\% improvement on $AP_{75}$. When compared to representative layer interaction models, DLA-L continues to exhibit excellence. Despite introducing more parameters compared to RLA$_g$ \cite{zhao2021recurrence}, there is an excellent increase in AP by 1.8\% and 1.1\% on DLA-L-50 and DLA-L-101, respectively. Meanwhile, in contrast to a static layer attention model, MRLA-L \cite{fang2023cross}, our DLA-L achieves a respective increase of 0.2\% and 0.3\% in AP. When utilizing Mask R-CNN as the detector, our DLA-L also outperforms the aforementioned models. Additionally, it surpasses NL \cite{wang2018non}, GC \cite{cao2019gcnet}, BA \cite{zhao2022ba}, and other models, showcasing the remarkable potential of the DLA architecture in facilitating dynamic modification in inter-layer information, even with the introduction of a tolerable parameter increment.

\section{Ablation Study}
\subsection{Evaluating Different RNN Blocks in DLA-L}

In order to validate the effectiveness of our proposed DSU block in implementing dynamic layer attention, we incorporated the original RNN block, DIA block and LSTM block as plugins into DLA to replace DSU block for comparative experiments. Due to limitations in computational resources, our ablation experiments were conducted using ResNet-110 as the baseline on CIFAR-100. Each experiment was run five times, and the results were expressed in the form of Accuracy $\pm$ Std.

\begin{table}[ht]
\centering
\setlength{\tabcolsep}{5mm}{
\begin{tabular}{lcc}
\toprule
\textbf{Block} & \textbf{Params} & \textbf{Top-1 acc. (\%)} \\
\midrule
Original RNN & 1.41M & 73.65$\pm$0.23 \\
LSTM & 1.93M & 75.32$\pm$0.34 \\
DIA & 1.46M & 75.60$\pm$0.36 \\
DSU & 1.39M & \textbf{76.41$\pm$0.36}  \\
\bottomrule
\end{tabular}}
\caption{Testing accuracy of different RNN blocks in DLA-110 on the CIFAR-100 dataset. }\label{tab:4}
\end{table}

As illustrated in Table \ref{tab:4}, the dynamic layer attention implemented by the four RNN blocks exhibits varying performance. The original RNN blocks show the poorest performance, with even a little top-1 accuracy increase than the baseline. LSTM block and the DIA block demonstrate comparable performance, outperforming baseline by 2.32\% and 2.60\%, respectively. However, the LSTM block has an additional parameter count of 0.47M compared to the DIA block, making it impractical applying LSTM in DLA. On the other hand, our proposed DSU block, utilizing even fewer parameters, achieves superior results. In comparison to the DIA block, we have 0.07M fewer parameters, leading to a 0.81\% improvement in top-1 accuracy. Therefore, it can be inferred that our proposed DSU is the most effective block among existing RNN blocks for implementing DLA.

\subsection{\texorpdfstring{Evaluating the Introduced $\sigma$ Function in DSU}{Evaluating the Introduced sigma Function in DSU}}

Compared to other RNN blocks, our DSU block first normalizes \(c^{m-1}\) using an activation function, where we employ the Sigmoid function \(\sigma(z) = \frac{1}{1 + e^{-z}}\) for this purpose. To contrast the role of the \(\sigma\) function utilized in DSU, we will substitute it with various activation functions, including Identity mapping, Tanh, and ReLU functions.

\begin{table}[htbp]
  \centering
  
  \setlength{\tabcolsep}{3.5mm}{
    \begin{tabular}{c|lc}
    \hline
    
    \hline
         \textbf{Model} & \textbf{Function} & \multicolumn{1}{l}{\textbf{Top-1 acc. (\%)}} \bigstrut\\
    \hline
    
    \hline
    \multirow{4}[2]{*}{DLA-L-56} & Identity mapping  & 74.63$\pm$0.14 \bigstrut[t]\\
          & Tanh &  74.88$\pm$0.21 \\
          & ReLU & 74.95$\pm$0.40 \\
          & Sigmoid & \textbf{75.46$\pm$0.26} \\
    \hline
    \multirow{4}[2]{*}{DLA-L-110} & Identity mapping  & 75.37$\pm$0.38  \bigstrut[t]\\
         
          & Tanh &   75.67$\pm$0.20 \\
          & ReLU     & 75.99$\pm$0.29 \\
          & Sigmoid & \textbf{76.41$\pm$0.36} \bigstrut[b]\\
    \hline
    
    \hline
    \end{tabular}}
    \caption{Testing accuracy of different activation functions used for normalizing \(c^{m-1}\) in DLA-L on CIFAR-100 dataset.}
  \label{tab:5}
\end{table}

As shown in Table \ref{tab:5}, we conducted experiments on the CIFAR-100 dataset using DLA-L-56 and DLA-L-110. Firstly, we observed in DSU that normalizing \(c^{m-1}\) is essential compared to Identity Mapping. The inclusion of an activation function resulted in significantly higher top-1 accuracy compared to Identity mapping. Secondly, among various activation functions, the Sigmoid function has been proven effective in scaling \(c^{m-1}\) to the range of $(0, 1)$, facilitating better fusion with \(y^m\) as input for the DSU block. Experimental results confirm this observation, achieving impressive top-1 accuracy of 75.46\% and 76.41\% on DLA-L-56 and DLA-L-110, respectively, far surpassing the baseline.

\subsection{Evaluating DSU Block in Layer-wise Information Integration}
We also attempt to evaluate that whether the proposed DSU block could be deployed to other layer interaction methods. \cite{huang2020dianet} introduced a Layer-wise Information Integration (LII) architecture that shared a RNN block throughout different network layers. And DIANet serves as a simple and effective form of LII architecture, demonstrating notable achievements in image classification. 
We evaluated the performance of the proposed DSU, LSTM, and DIA blocks, when integrating them into the LII. Additionally, we employed ResNets as the backbone and conduct experimental comparisons on the CIFAR-100 dataset.

\begin{table}[ht]
\centering
\setlength{\tabcolsep}{6mm}{
\begin{tabular}{lcc}
\toprule
\textbf{Block} & \textbf{Params} & \textbf{Top-1 acc. (\%)} \\
\midrule
ResNet-56 & 0.61M & 72.32$\pm$0.36 \\
LII (LSTM) & 1.31M    & 69.28$\pm$0.44    \\
LII (DIA) & 0.83M & 73.87$\pm$0.27 \\
LII (DSU) & 0.79M & \textbf{74.23$\pm$0.22} \\

\midrule
ResNet-110 & 1.17M & 73.00$\pm$0.36 \\
LII (LSTM) &  1.86M   & 71.31$\pm$0.33   \\
LII (DIA) & 1.39M & \textbf{75.31$\pm$0.16} \\
LII (DSU) & 1.35M & 75.14$\pm$0.21 \\

\bottomrule
\end{tabular}}
\caption{Testing accuracy of different RNN blocks in layer-wise information integration on the CIFAR-100 dataset. }
\label{tab:6}
\end{table}

As shown in Table \ref{tab:6}, the experimental results demonstrate that the RNN blocks exhibits favorable performance when applied to LII. On the CIFAR-100 dataset, our DSU block shows improvements of 1.91\% and 2.14\% on ResNet-56 and ResNet-110, respectively. While the introduced parameters are 0.04M smaller than DIA block, DSU block outperforms DIA block by 0.36\% on ResNet-56. Meanwhile, DSU block performs comparably with DIA block on ResNet-110. On the contrary, LSTM block has the highest number of parameters but performs the worst. Therefore, it could be concluded that DSU block is also a challenging method in LII, which could achieve comparable results to DIA block in LII architecture while having fewer parameters.

\section{Conclusion}

In this paper, we first unveil the inherent static nature of existing layer attention mechanisms and analyze their limitations in facilitating feature extraction through layer interaction. To address these limitations and restore the dynamic features of attention mechanisms, we propose and construct a framework called Dynamic Layer Attention (DLA). For implementing DLA, we design a novel RNN block, named Dynamic Sharing Unit (DSU). Experimental evaluations in the domains of image classification and object detection demonstrate that our framework outperforms static layer attention significantly in promoting layer interaction.

\section*{Acknowledgments}
This work was supported by the National Major Science and
Technology Projects of China under Grant 2018AAA0100201, the National Natural Science Foundation of China under Grant 62206189, and the China Postdoctoral Science Foundation under Grant 2023M732427.

\bibliographystyle{named}
\bibliography{ijcai24}

\begin{thebibliography}{}

\bibitem[\protect\citeauthoryear{Bello \bgroup \em et al.\egroup }{2019}]{bello2019attention}
Irwan Bello, Barret Zoph, Ashish Vaswani, Jonathon Shlens, and Quoc~V Le.
\newblock Attention augmented convolutional networks.
\newblock In {\em ICCV}, pages 3286--3295, 2019.

\bibitem[\protect\citeauthoryear{Cao \bgroup \em et al.\egroup }{2019}]{cao2019gcnet}
Yue Cao, Jiarui Xu, Stephen Lin, Fangyun Wei, and Han Hu.
\newblock Gcnet: Non-local networks meet squeeze-excitation networks and beyond.
\newblock In {\em ICCV}, pages 0--0, 2019.

\bibitem[\protect\citeauthoryear{Carion \bgroup \em et al.\egroup }{2020}]{carion2020end}
Nicolas Carion, Francisco Massa, Gabriel Synnaeve, Nicolas Usunier, Alexander Kirillov, and Sergey Zagoruyko.
\newblock End-to-end object detection with transformers.
\newblock In {\em ECCV}, pages 213--229, 2020.

\bibitem[\protect\citeauthoryear{Chen \bgroup \em et al.\egroup }{2018a}]{chen2018video}
Dapeng Chen, Hongsheng Li, Tong Xiao, Shuai Yi, and Xiaogang Wang.
\newblock Video person re-identification with competitive snippet-similarity aggregation and co-attentive snippet embedding.
\newblock In {\em CVPR}, pages 1169--1178, 2018.

\bibitem[\protect\citeauthoryear{Chen \bgroup \em et al.\egroup }{2018b}]{chen20182}
Yunpeng Chen, Yannis Kalantidis, Jianshu Li, Shuicheng Yan, and Jiashi Feng.
\newblock A\^{} 2-nets: Double attention networks.
\newblock {\em NeurIPS}, 2018.

\bibitem[\protect\citeauthoryear{Chen \bgroup \em et al.\egroup }{2019}]{chen2019mmdetection}
Kai Chen, Jiaqi Wang, Jiangmiao Pang, Yuhang Cao, Yu~Xiong, Xiaoxiao Li, Shuyang Sun, Wansen Feng, Ziwei Liu, Jiarui Xu, et~al.
\newblock Mmdetection: Open mmlab detection toolbox and benchmark.
\newblock {\em arXiv preprint arXiv:1906.07155}, 2019.

\bibitem[\protect\citeauthoryear{Chen \bgroup \em et al.\egroup }{2023}]{chen2023dual}
Zheng Chen, Yulun Zhang, Jinjin Gu, Linghe Kong, Xiaokang Yang, and Fisher Yu.
\newblock Dual aggregation transformer for image super-resolution.
\newblock In {\em ICCV}, pages 12312--12321, 2023.

\bibitem[\protect\citeauthoryear{Fang \bgroup \em et al.\egroup }{2023}]{fang2023cross}
Yanwen Fang, Yuxi Cai, Jintai Chen, Jingyu Zhao, Guangjian Tian, and Guodong Li.
\newblock Cross-layer retrospective retrieving via layer attention.
\newblock {\em arXiv preprint arXiv:2302.03985}, 2023.

\bibitem[\protect\citeauthoryear{Han \bgroup \em et al.\egroup }{2021}]{han2021dynamic}
Yizeng Han, Gao Huang, Shiji Song, Le~Yang, Honghui Wang, and Yulin Wang.
\newblock Dynamic neural networks: A survey.
\newblock {\em IEEE Transactions on Pattern Analysis and Machine Intelligence}, pages 7436--7456, 2021.

\bibitem[\protect\citeauthoryear{He \bgroup \em et al.\egroup }{2016}]{he2016deep}
Kaiming He, Xiangyu Zhang, Shaoqing Ren, and Jian Sun.
\newblock Deep residual learning for image recognition.
\newblock In {\em CVPR}, pages 770--778, 2016.

\bibitem[\protect\citeauthoryear{He \bgroup \em et al.\egroup }{2017}]{he2017mask}
Kaiming He, Georgia Gkioxari, Piotr Doll{\'a}r, and Ross Girshick.
\newblock Mask r-cnn.
\newblock In {\em ICCV}, pages 2961--2969, 2017.

\bibitem[\protect\citeauthoryear{He \bgroup \em et al.\egroup }{2019}]{he2019bag}
Tong He, Zhi Zhang, Hang Zhang, Zhongyue Zhang, Junyuan Xie, and Mu~Li.
\newblock Bag of tricks for image classification with convolutional neural networks.
\newblock In {\em CVPR}, pages 558--567, 2019.

\bibitem[\protect\citeauthoryear{He \bgroup \em et al.\egroup }{2020}]{he2020realformer}
Ruining He, Anirudh Ravula, Bhargav Kanagal, and Joshua Ainslie.
\newblock Realformer: Transformer likes residual attention.
\newblock {\em arXiv preprint arXiv:2012.11747}, 2020.

\bibitem[\protect\citeauthoryear{Hochreiter and Schmidhuber}{1997}]{hochreiter1997long}
Sepp Hochreiter and J{\"u}rgen Schmidhuber.
\newblock Long short-term memory.
\newblock {\em Neural Computation}, pages 1735--1780, 1997.

\bibitem[\protect\citeauthoryear{Hu \bgroup \em et al.\egroup }{2018}]{hu2018squeeze}
Jie Hu, Li~Shen, and Gang Sun.
\newblock Squeeze-and-excitation networks.
\newblock In {\em CVPR}, pages 7132--7141, 2018.

\bibitem[\protect\citeauthoryear{Huang \bgroup \em et al.\egroup }{2016}]{huang2016deep}
Gao Huang, Yu~Sun, Zhuang Liu, Daniel Sedra, and Kilian~Q Weinberger.
\newblock Deep networks with stochastic depth.
\newblock In {\em ECCV}, pages 646--661, 2016.

\bibitem[\protect\citeauthoryear{Huang \bgroup \em et al.\egroup }{2017}]{huang2017densely}
Gao Huang, Zhuang Liu, Laurens Van Der~Maaten, and Kilian~Q Weinberger.
\newblock Densely connected convolutional networks.
\newblock In {\em CVPR}, pages 4700--4708, 2017.

\bibitem[\protect\citeauthoryear{Huang \bgroup \em et al.\egroup }{2018}]{huang2018condensenet}
Gao Huang, Shichen Liu, Laurens Van~der Maaten, and Kilian~Q Weinberger.
\newblock Condensenet: An efficient densenet using learned group convolutions.
\newblock In {\em CVPR}, pages 2752--2761, 2018.

\bibitem[\protect\citeauthoryear{Huang \bgroup \em et al.\egroup }{2019}]{huang2019ccnet}
Zilong Huang, Xinggang Wang, Lichao Huang, Chang Huang, Yunchao Wei, and Wenyu Liu.
\newblock Ccnet: Criss-cross attention for semantic segmentation.
\newblock In {\em ICCV}, pages 603--612, 2019.

\bibitem[\protect\citeauthoryear{Huang \bgroup \em et al.\egroup }{2020}]{huang2020dianet}
Zhongzhan Huang, Senwei Liang, Mingfu Liang, and Haizhao Yang.
\newblock Dianet: Dense-and-implicit attention network.
\newblock In {\em AAAI}, pages 4206--4214, 2020.

\bibitem[\protect\citeauthoryear{Jiao \bgroup \em et al.\egroup }{2023}]{jiao2023dilateformer}
Jiayu Jiao, Yu-Ming Tang, Kun-Yu Lin, Yipeng Gao, Jinhua Ma, Yaowei Wang, and Wei-Shi Zheng.
\newblock Dilateformer: Multi-scale dilated transformer for visual recognition.
\newblock {\em IEEE Transactions on Multimedia}, 2023.

\bibitem[\protect\citeauthoryear{Li and Chen}{2020}]{li2020deep}
Duo Li and Qifeng Chen.
\newblock Deep reinforced attention learning for quality-aware visual recognition.
\newblock In {\em ECCV}, pages 493--509, 2020.

\bibitem[\protect\citeauthoryear{Li \bgroup \em et al.\egroup }{2019}]{li2019selective}
Xiang Li, Wenhai Wang, Xiaolin Hu, and Jian Yang.
\newblock Selective kernel networks.
\newblock In {\em CVPR}, pages 510--519, 2019.

\bibitem[\protect\citeauthoryear{Liu \bgroup \em et al.\egroup }{2021}]{liu2021swin}
Ze~Liu, Yutong Lin, Yue Cao, Han Hu, Yixuan Wei, Zheng Zhang, Stephen Lin, and Baining Guo.
\newblock Swin transformer: Hierarchical vision transformer using shifted windows.
\newblock In {\em ICCV}, pages 10012--10022, 2021.

\bibitem[\protect\citeauthoryear{Ren \bgroup \em et al.\egroup }{2015}]{ren2015faster}
Shaoqing Ren, Kaiming He, Ross Girshick, and Jian Sun.
\newblock Faster r-cnn: Towards real-time object detection with region proposal networks.
\newblock {\em NeurIPS}, 2015.

\bibitem[\protect\citeauthoryear{Ronneberger \bgroup \em et al.\egroup }{2015}]{ronneberger2015u}
Olaf Ronneberger, Philipp Fischer, and Thomas Brox.
\newblock U-net: Convolutional networks for biomedical image segmentation.
\newblock In {\em MICCAI}, pages 234--241, 2015.

\bibitem[\protect\citeauthoryear{Srivastava \bgroup \em et al.\egroup }{2015}]{srivastava2015training}
Rupesh~K Srivastava, Klaus Greff, and J{\"u}rgen Schmidhuber.
\newblock Training very deep networks.
\newblock {\em NeurIPS}, 2015.

\bibitem[\protect\citeauthoryear{Vaswani \bgroup \em et al.\egroup }{2017}]{vaswani2017attention}
Ashish Vaswani, Noam Shazeer, Niki Parmar, Jakob Uszkoreit, Llion Jones, Aidan~N Gomez, {\L}ukasz Kaiser, and Illia Polosukhin.
\newblock Attention is all you need.
\newblock {\em NeurIPS}, 2017.

\bibitem[\protect\citeauthoryear{Wang \bgroup \em et al.\egroup }{2018}]{wang2018non}
Xiaolong Wang, Ross Girshick, Abhinav Gupta, and Kaiming He.
\newblock Non-local neural networks.
\newblock In {\em CVPR}, pages 7794--7803, 2018.

\bibitem[\protect\citeauthoryear{Wang \bgroup \em et al.\egroup }{2020a}]{wang2020eca}
Qilong Wang, Banggu Wu, Pengfei Zhu, Peihua Li, Wangmeng Zuo, and Qinghua Hu.
\newblock Eca-net: Efficient channel attention for deep convolutional neural networks.
\newblock In {\em CVPR}, pages 11534--11542, 2020.

\bibitem[\protect\citeauthoryear{Wang \bgroup \em et al.\egroup }{2020b}]{wang2020glance}
Yulin Wang, Kangchen Lv, Rui Huang, Shiji Song, Le~Yang, and Gao Huang.
\newblock Glance and focus: a dynamic approach to reducing spatial redundancy in image classification.
\newblock {\em NeurIPS}, pages 2432--2444, 2020.

\bibitem[\protect\citeauthoryear{Woo \bgroup \em et al.\egroup }{2018}]{woo2018cbam}
Sanghyun Woo, Jongchan Park, Joon-Young Lee, and In~So Kweon.
\newblock Cbam: Convolutional block attention module.
\newblock In {\em ECCV}, pages 3--19, 2018.

\bibitem[\protect\citeauthoryear{Xu \bgroup \em et al.\egroup }{2017}]{xu2017jointly}
Shuangjie Xu, Yu~Cheng, Kang Gu, Yang Yang, Shiyu Chang, and Pan Zhou.
\newblock Jointly attentive spatial-temporal pooling networks for video-based person re-identification.
\newblock In {\em ICCV}, pages 4733--4742, 2017.

\bibitem[\protect\citeauthoryear{Yang \bgroup \em et al.\egroup }{2020}]{yang2020resolution}
Le~Yang, Yizeng Han, Xi~Chen, Shiji Song, Jifeng Dai, and Gao Huang.
\newblock Resolution adaptive networks for efficient inference.
\newblock In {\em CVPR}, pages 2369--2378, 2020.

\bibitem[\protect\citeauthoryear{Yang \bgroup \em et al.\egroup }{2021}]{yang2021condensenet}
Le~Yang, Haojun Jiang, Ruojin Cai, Yulin Wang, Shiji Song, Gao Huang, and Qi~Tian.
\newblock Condensenet v2: Sparse feature reactivation for deep networks.
\newblock In {\em CVPR}, pages 3569--3578, 2021.

\bibitem[\protect\citeauthoryear{Zhao \bgroup \em et al.\egroup }{2021}]{zhao2021recurrence}
Jingyu Zhao, Yanwen Fang, and Guodong Li.
\newblock Recurrence along depth: Deep convolutional neural networks with recurrent layer aggregation.
\newblock {\em NeurIPS}, pages 10627--10640, 2021.

\bibitem[\protect\citeauthoryear{Zhao \bgroup \em et al.\egroup }{2022}]{zhao2022ba}
Yue Zhao, Junzhou Chen, Zirui Zhang, and Ronghui Zhang.
\newblock Ba-net: Bridge attention for deep convolutional neural networks.
\newblock In {\em ECCV}, pages 297--312, 2022.

\end{thebibliography}

\end{document}